\documentclass{article}
\usepackage{spconf,amsmath,epsfig}
\usepackage{breqn}
\usepackage{dblfloatfix}    
\graphicspath{{Figures/}}
\usepackage{graphicx}
\usepackage{lipsum}
\usepackage{url}
\usepackage{color}
\usepackage{bm}
\usepackage{balance}
\usepackage{tabto}
\usepackage[table]{xcolor}
\usepackage{enumitem}

\usepackage{array}
\newcommand{\PreserveBackslash}[1]{\let\temp=\\#1\let\\=\temp}
\newcolumntype{C}[1]{>{\PreserveBackslash\centering}p{#1}}
\newcolumntype{R}[1]{>{\PreserveBackslash\raggedleft}p{#1}}
\newcolumntype{L}[1]{>{\PreserveBackslash\raggedright}p{#1}}

\title{Automated Coastline Extraction using Edge Detection Algorithms}
\name{Conor~O'Sullivan$^{1,2}$, Seamus~Coveney$^{3}$, Xavier~Monteys$^{4}$, Soumyabrata Dev$^{1,2}$ 
\thanks{This publication has emanated from research conducted with the financial support of Science Foundation Ireland under Grant number 18/CRT/6183. This research was conducted with the financial support of Science Foundation Ireland under Grant Agreement No.\ 13/RC/2106\_P2 at the ADAPT SFI Research Centre at University College Dublin. ADAPT, the SFI Research Centre for AI-Driven Digital Content Technology, is funded by Science Foundation Ireland through the SFI Research Centres Programme.}
\thanks{Send correspondence to S.\ Dev: \protect\url{soumyabrata.dev@ucd.ie}}
}

\address{
$^{1}$The ADAPT SFI Research Centre, Dublin, Ireland \\
$^{2}$School of Computer Science, University College Dublin, Ireland\\
$^{3}$Envo-Geo Environmental Geoinformatics, Skibbereen, Ireland \\
$^{4}$Geological Survey Ireland, Dublin, Ireland
}
\begin{document}
%
\maketitle
\begin{abstract}
We analyse the effectiveness of edge detection algorithms for the purpose of automatically extracting coastlines from satellite images. Four algorithms - Canny, Sobel, Scharr and Prewitt are compared visually and using metrics. With an average SSIM of 0.8, Canny detected edges that were closest to the reference edges. However, the algorithm had difficulty distinguishing noisy edges, e.g. due to development, from coastline edges. In addition, histogram equalization and Gaussian blur were shown to improve the effectiveness of the edge detection algorithms by up to 1.5 and 1.6 times respectively. 
\end{abstract}

\vspace{-0.5cm}
\section{Introduction}

Humans depend on the coast. At the same time, it is a dynamic environment. It can be impacted by erosion, sedimentation and human activities such as land development. In fact, 37\% of coastlines classified as nature protected areas are already experiencing erosion~\cite{luijendijk2018state}. To make matters worse, climate change will likely accelerate this process~\cite{ranasinghe2016assessing}. These hazards can have negative consequences for the communities, economies and ecosystems that rely on the coastline. 

These consequences warrant the monitoring of coastlines. This is to understand these hazards and how to best mitigate the risks involved. Yet, with an estimated 1,634,701 km of global coastline, this is no simple task~\cite{burke2001coastal}. To monitor the coastline effectively at scale requires the use of satellite images and automated methods~\cite{osullivan2023interpreting}. In this paper, we analyse the effectiveness of edge detection algorithms for automated coastline detection.

\subsection{Related work}
Edge detection algorithms are fast and do not require training data. The downside is they are not robust to noise in satellite images caused by factors like cloud cover and development. In addition, research has focused on applying these approaches to small datasets of individual coastlines such as in Dubai~\cite{vukadinov2017algorithm}, Antarctica~\cite{klinger2011antarctic} and Greece~\cite{paravolidakis2018automatic}. It is not clear how well these algorithms can perform across a variety of coastline types. 

As a more robust solution, deep learning methods have been proposed. Here the problem is framed as an image segmentation task. Models predict a binary map that separates land from water. Variations of U-Net have been shown to perform well for similar remote sensing tasks such as cloud segmentation~\cite{guo2020cloud}. For coastline extraction, deep learning methods have accurately segmented five coastlines across three continents~\cite{vos2019sub}. 

Recently, the Sentinel-2 Water Edges Dataset (SWED) was released~\cite{seale2022swed}. With 49 testing locations, this is the most diverse dataset for coastline extraction. The researchers showed that a variation of U-Net was able to accurately segment across all coastline types. When training these models, all 12 available sentinel-2 bands were used and only the cost function was varied. It is not clear how preprocessing the satellite bands may aid predictions or which bands were most useful for predictions. 

\vspace{-0.3cm}
\subsection{Contributions of this paper}
The introduction of SWED provides an opportunity to benchmark automated approaches across a diverse dataset of coastlines. Our main contributions are:

\vspace{-0.2cm}
\begin{itemize}
    \item Analyse the effectiveness of 4 edge detection algorithms to detect coastlines.
    \vspace{-0.2cm}
    \item Providing insight into feature engineering methods that may benefit deep learning methods.
    \vspace{-0.2cm}
    \item We open source the code to reproduce the results of this paper. This can be found in the GitHub repository~\footnote{In the spirt of reproducible research, the codes to reproduce the results of this paper can be found here: \url{https://github.com/conorosully/SWED-edge-detection}.}.
\end{itemize}

\section{Methodology}

\subsection{Dataset}

\begin{figure}[h]
\centering
\includegraphics[width=0.48\textwidth]{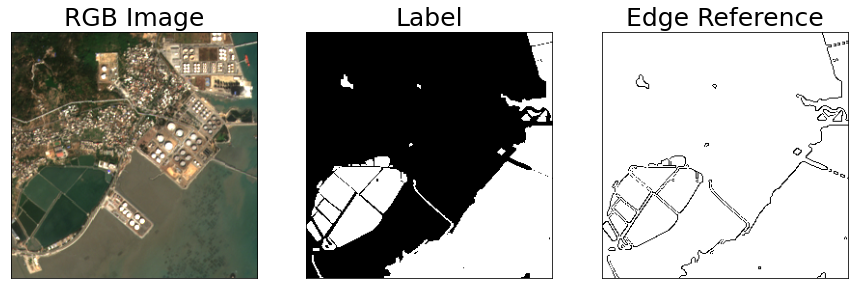}
\caption{Example of a test image, binary land/water map, and coastline edge obtained from SWED~\cite{seale2022swed}.}
\label{fig:example}
\end{figure}

We use 98 test images and labels from SWED~\cite{seale2022swed} as a reference dataset. We do not consider the training set as it has been labeled using semi-supervised methods. In comparison, more care has been taken when labeling the test set. We demonstrate an example of an image and its corresponding binary label map and edge reference in Figure~\ref{fig:example}. The label is a binary segmentation map. Each pixel is either classified as land (black) or water (white). To reformulate this as an edge detection problem, we convert these labels to edge maps. This is done using the canny edge detection algorithm. These edge maps are taken as the reference and used to compare the output of the edge detection algorithms. 

As input, 12 spectral bands are available from sentinel-2 imagery~\cite{osullivan2023analyzing}. They include visible red, green and blue wavelengths. The example image in Figure~\ref{fig:example} has been created by combining these three bands. The bands also include shortwave inferred wavelengths. The spatial resolution of the bands varies between 10m, 20m and 60m. We considered each of the bands separately and treat them as grayscale images.

\subsection{Pre-processing}
Before applying the algorithms, the bands are pre-processed by:

\begin{enumerate}[noitemsep]
    \item Scaling the pixel values between 0 and 255 
    \item Increasing the contrast using Equalized Histogram
    \item Reducing noise using Gaussian blur 
\end{enumerate}

\begin{figure}[h]
\centering
\includegraphics[width=0.45\textwidth]{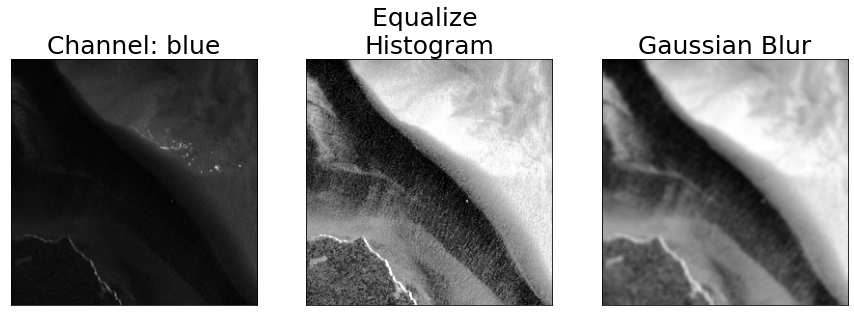}
\caption{Example of pre-processing steps using blue band as an example.}
\label{fig:preprocessing}
\end{figure}

In Figure~\ref{fig:preprocessing}, you can see an example of this process. We use the blue band as an example.  This is because, before the processing is applied, the pixel values for this band tend to be skewed towards less intense values. In other words, the increased contrast due to step 2 is clear. Visually, the effect of the processing was similar for the remaining bands. 

\subsection{Edge Detection and Metrics}
We apply common edge detection algorithms - Canny, Sobel, Scharr and Prewitt. The algorithms are applied to each of the 12 bands individually. We then compare the results to the edge map references. The comparison is done visually and using 4 different metrics - root mean square error (RMSE), peak signal-to-noise ratio (PSNR), universal image quality index (UQI) and structural similarity (SSIM). The metrics give us a quantification of how well the algorithms detect coastal edges when compared to the reference edges. We take the average of these metrics across the 98 test images. We also calculate error bars by taking the standard deviation. 

\section{Results \& Discussion}
\subsection{Subjective visual analysis}
\begin{figure}[h]
\centering
\includegraphics[width=0.45\textwidth]{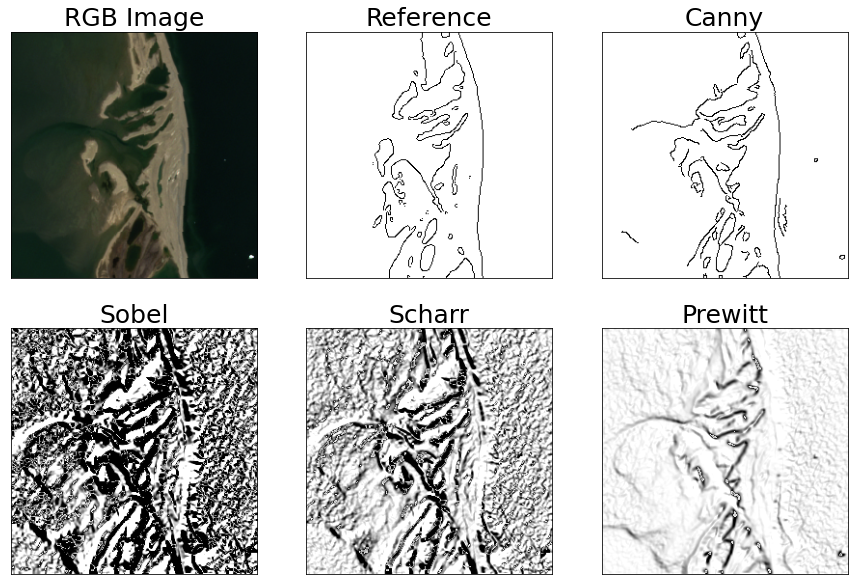}
\caption{We demonstrate the results obtained from edge detection algorithms applied to the blue band. We observe Canny to provide the best edge map followed by Prewitt.}
\label{fig:edge}
\end{figure}

When comparing the algorithms of the methods visually, Canny produced the best results. In general, less noisy non-coastal edges were detected. In Figure~\ref{fig:edge}, we can see the output of all 4 methods when applied to one of the test images. Again, we have used the blue band as an example. This band has a high spatial resolution (10m) and the results can be visually verified. The results were similar for the remaining bands. 

In Figure~\ref{fig:edge}, we can see that Sobel and Scharr have detected many edges apart from the coastline. This is potentially due to factors like wind and swells causing ripples in the ocean. Prewitt has also detected some noisy edges but the results are visually better. Canny was able to identify most of the relevant edges. Yet, we can see the method still missed some coastal edges and identified some non-coastal edges. 

\begin{figure}[h]
\centering
\includegraphics[width=0.48\textwidth]{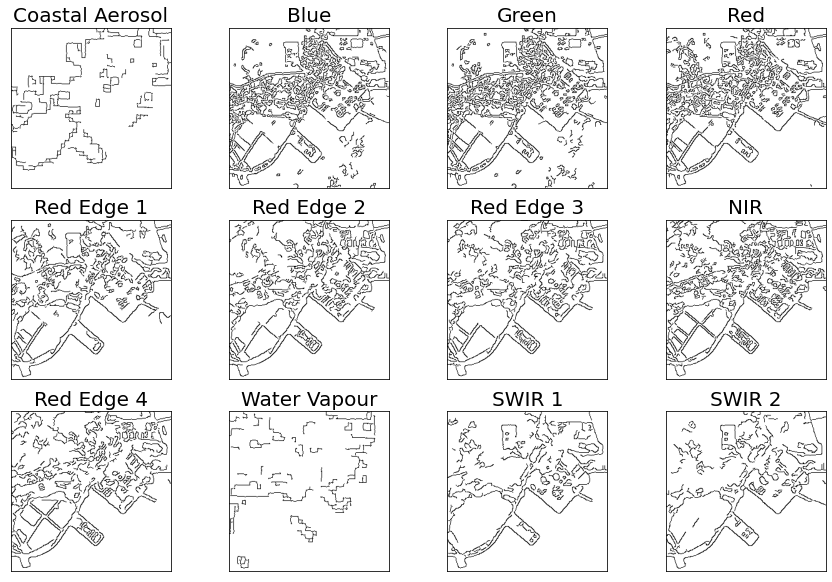}
\caption{We apply Canny edge detection to all channels of a SWED image.}
\label{fig:details_channels}
\end{figure}

Although Canny appeared to perform well in some locations, it was not robust to development on the coastline. For example, take the image in Figure~\ref{fig:example}. We can see that buildings on the coastline create non-coastal edges. In Figure~\ref{fig:details_channels}, we can see that, across all 12 bands, Canny fails to distinguish these from water edges. This suggests that edge detection methods are not appropriate when dealing with multiple types of coastlines. 

\subsection{Objective evaluation metrics}

The metrics confirmed the visual analysis. In Table~\ref{tab:main_results}, we have the PSNR and SSIM metrics for all algorithms and bands. When using PSNR, Prewitt had the best performance across all the bands and Canny was a close second. When using SSIM, Canny had the best performance across all 12 bands. 

In Table~\ref{tab:main_results} the Coastal Aerosol and Water Vapour bands had the best performance. Specifically, the bands had the highest PSNR and SSIM values across all 4 edge detection algorithms. We should consider that this is potentially due to the resolution of these bands which is 60m. In comparison, the remaining bands had a resolution of either 20m or 10m. Due to the lower resolution, noisy edges may be less prominent in the Coastal Aerosol and Water Vapour bands. 

We do not give the results for RMSE and UQI due to space constraints. In addition, RMSE produced similar results to PSNR. This is expected as its calculation is related to PSNR. Similarly, the calculation of UQI is related to SSIM and provided similar results. 

\begin{table*}[t]
\centering
\begin{tabular}{c|ll|ll|ll|ll|}
\cline{2-9}
                                                              & \multicolumn{2}{c|}{\textbf{Canny}}                                                          & \multicolumn{2}{c|}{\textbf{Sobel}}                                                         & \multicolumn{2}{c|}{\textbf{Scharr}}                                                        & \multicolumn{2}{c|}{\textbf{Prewitt}}                                                        \\ \hline
\multicolumn{1}{|c|}{\textbf{Band}}                           & \multicolumn{1}{c|}{\textbf{PSNR}}                      & \multicolumn{1}{c|}{\textbf{SSIM}} & \multicolumn{1}{c|}{\textbf{PSNR}}                     & \multicolumn{1}{c|}{\textbf{SSIM}} & \multicolumn{1}{c|}{\textbf{PSNR}}                     & \multicolumn{1}{c|}{\textbf{SSIM}} & \multicolumn{1}{c|}{\textbf{PSNR}}                      & \multicolumn{1}{c|}{\textbf{SSIM}} \\ \hline
\rowcolor[HTML]{EFEFEF} 
\multicolumn{1}{|c|}{\cellcolor[HTML]{EFEFEF}Coastal Aerosol} & \multicolumn{1}{l|}{\cellcolor[HTML]{EFEFEF}13.2 ± 2}   & 0.8 ± 0.1                          & \multicolumn{1}{l|}{\cellcolor[HTML]{EFEFEF}4.3 ± 0.8} & 0.2 ± 0.1                          & \multicolumn{1}{l|}{\cellcolor[HTML]{EFEFEF}8.1 ± 1.3} & 0.2 ± 0.1                          & \multicolumn{1}{l|}{\cellcolor[HTML]{EFEFEF}14.5 ± 1.7} & 0.3 ± 0.1                          \\ \hline
\multicolumn{1}{|c|}{Blue}                                    & \multicolumn{1}{l|}{10.3 ± 2.4}                         & 0.6 ± 0.2                          & \multicolumn{1}{l|}{3 ± 0.5}                           & 0 ± 0                              & \multicolumn{1}{l|}{5.6 ± 1.1}                         & 0 ± 0                              & \multicolumn{1}{l|}{12.6 ± 1.7}                         & 0.1 ± 0                            \\ \hline
\rowcolor[HTML]{EFEFEF} 
\multicolumn{1}{|c|}{\cellcolor[HTML]{EFEFEF}Green}           & \multicolumn{1}{l|}{\cellcolor[HTML]{EFEFEF}10.8 ± 2.5} & 0.7 ± 0.2                          & \multicolumn{1}{l|}{\cellcolor[HTML]{EFEFEF}3.1 ± 0.6} & 0 ± 0                              & \multicolumn{1}{l|}{\cellcolor[HTML]{EFEFEF}6 ± 1.2}   & 0 ± 0                              & \multicolumn{1}{l|}{\cellcolor[HTML]{EFEFEF}13 ± 1.9}   & 0.1 ± 0.1                          \\ \hline
\multicolumn{1}{|c|}{Red}                                     & \multicolumn{1}{l|}{10.9 ± 2.7}                         & 0.7 ± 0.2                          & \multicolumn{1}{l|}{3.3 ± 0.9}                         & 0.1 ± 0.1                          & \multicolumn{1}{l|}{6.1 ± 1.4}                         & 0.1 ± 0.1                          & \multicolumn{1}{l|}{13.1 ± 2.1}                         & 0.1 ± 0.1                          \\ \hline
\rowcolor[HTML]{EFEFEF} 
\multicolumn{1}{|c|}{\cellcolor[HTML]{EFEFEF}Red Edge 1}      & \multicolumn{1}{l|}{\cellcolor[HTML]{EFEFEF}11.5 ± 2.7} & 0.7 ± 0.2                          & \multicolumn{1}{l|}{\cellcolor[HTML]{EFEFEF}3.4 ± 0.9} & 0.1 ± 0.1                          & \multicolumn{1}{l|}{\cellcolor[HTML]{EFEFEF}6.4 ± 1.5} & 0.1 ± 0.1                          & \multicolumn{1}{l|}{\cellcolor[HTML]{EFEFEF}13.4 ± 2}   & 0.2 ± 0.1                          \\ \hline
\multicolumn{1}{|c|}{Red Edge 2}                              & \multicolumn{1}{l|}{11.1 ± 2.4}                         & 0.7 ± 0.2                          & \multicolumn{1}{l|}{3.5 ± 0.9}                         & 0.1 ± 0.1                          & \multicolumn{1}{l|}{6.3 ± 1.3}                         & 0.1 ± 0.1                          & \multicolumn{1}{l|}{13.4 ± 1.7}                         & 0.2 ± 0.1                          \\ \hline
\rowcolor[HTML]{EFEFEF} 
\multicolumn{1}{|c|}{\cellcolor[HTML]{EFEFEF}Red Edge 3}      & \multicolumn{1}{l|}{\cellcolor[HTML]{EFEFEF}11.2 ± 2.4} & 0.7 ± 0.2                          & \multicolumn{1}{l|}{\cellcolor[HTML]{EFEFEF}3.5 ± 0.9} & 0.1 ± 0.1                          & \multicolumn{1}{l|}{\cellcolor[HTML]{EFEFEF}6.4 ± 1.3} & 0.1 ± 0.1                          & \multicolumn{1}{l|}{\cellcolor[HTML]{EFEFEF}13.5 ± 1.8} & 0.2 ± 0.1                          \\ \hline
\multicolumn{1}{|c|}{NIR}                                     & \multicolumn{1}{l|}{10.9 ± 2.6}                         & 0.7 ± 0.2                          & \multicolumn{1}{l|}{3.5 ± 0.9}                         & 0.1 ± 0.1                          & \multicolumn{1}{l|}{6.3 ± 1.3}                         & 0.1 ± 0.1                          & \multicolumn{1}{l|}{13.5 ± 1.8}                         & 0.2 ± 0.1                          \\ \hline
\rowcolor[HTML]{EFEFEF} 
\multicolumn{1}{|c|}{\cellcolor[HTML]{EFEFEF}Red Edge 4}      & \multicolumn{1}{l|}{\cellcolor[HTML]{EFEFEF}10.8 ± 2.3} & 0.7 ± 0.2                          & \multicolumn{1}{l|}{\cellcolor[HTML]{EFEFEF}3.4 ± 0.9} & 0.1 ± 0.1                          & \multicolumn{1}{l|}{\cellcolor[HTML]{EFEFEF}6.3 ± 1.3} & 0.1 ± 0.1                          & \multicolumn{1}{l|}{\cellcolor[HTML]{EFEFEF}13.3 ± 1.8} & 0.2 ± 0.1                          \\ \hline
\multicolumn{1}{|c|}{Water Vapour}                            & \multicolumn{1}{l|}{13.1 ± 1.8}                         & 0.8 ± 0.1                          & \multicolumn{1}{l|}{4.9 ± 1.3}                         & 0.3 ± 0.2                          & \multicolumn{1}{l|}{8.4 ± 1.4}                         & 0.3 ± 0.1                          & \multicolumn{1}{l|}{14.6 ± 1.7}                         & 0.4 ± 0.1                          \\ \hline
\rowcolor[HTML]{EFEFEF} 
\multicolumn{1}{|c|}{\cellcolor[HTML]{EFEFEF}SWIR 1}          & \multicolumn{1}{l|}{\cellcolor[HTML]{EFEFEF}11.7 ± 2.5} & 0.7 ± 0.1                          & \multicolumn{1}{l|}{\cellcolor[HTML]{EFEFEF}3.6 ± 0.9} & 0.1 ± 0.1                          & \multicolumn{1}{l|}{\cellcolor[HTML]{EFEFEF}6.6 ± 1.3} & 0.1 ± 0.1                          & \multicolumn{1}{l|}{\cellcolor[HTML]{EFEFEF}13.7 ± 1.8} & 0.2 ± 0.1                          \\ \hline
\multicolumn{1}{|c|}{SWIR 2}                                  & \multicolumn{1}{l|}{10.7 ± 2.4}                         & 0.7 ± 0.2                          & \multicolumn{1}{l|}{3.2 ± 0.5}                         & 0.1 ± 0                            & \multicolumn{1}{l|}{6 ± 1}                             & 0.1 ± 0                            & \multicolumn{1}{l|}{13.2 ± 1.7}                         & 0.1 ± 0                            \\ \hline
\end{tabular}
\caption{PSNR and SSIM evaluation metrics for each edge detection algorithm and band. The table provides the average and standard deviation across the 98 images in the SWED test set. Using SSIM, Canny had the best performance. Using PSNR, Prewitt had the best performance with Canny as a close second. Out of the 12 bands, Coastal Aerosol had the best performance.}
\label{tab:main_results}
\vspace{-0.3cm}
\end{table*}

\subsection{Effect of pre-processing}
\begin{figure}[h]
\centering
\includegraphics[width=0.48\textwidth]{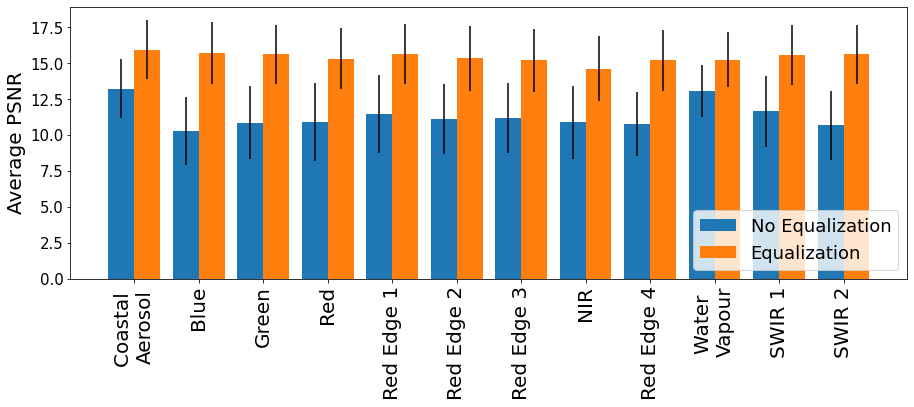}
\caption{Average PSNR for Canny edge detection with and without histogram equalization}
\label{fig:equalization}
\end{figure}

We investigated the effect of the preprocessing steps using PSNR and the Canny algorithm. In Figure~\ref{fig:equalization}, we have the effect of histogram equalization. We can see that it has improved the performance across all 12 bands. This suggests that increasing the contrast of the bands has made it easier to distinguish coastal edges. 

\begin{figure}[h]
\centering
\includegraphics[width=0.48\textwidth]{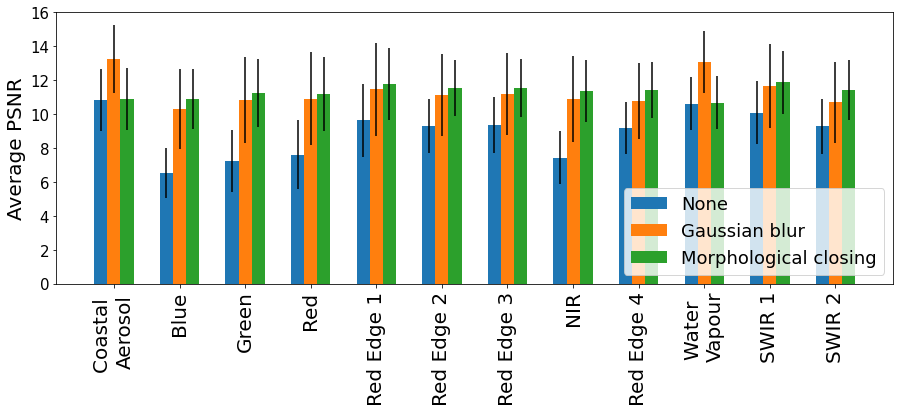}
\caption{Average PSNR for canny edge detection using different noise reduction methods}
\label{fig:noise}
\end{figure}

In Figure \ref{fig:noise}, we compare the effect of noise reduction methods. We include the results of an additional noise reduction method - morphological closing. 
We see that when no noise reduction is applied the algorithm tends to perform worse. This is compared to when either Gaussian blur or morphological closing is applied. 

For 10 of the bands, morphological closing appears to produce better results. However, we decided not to use this method in the main analysis. This is because the process of morphological closing creates new noisy edges in the images. In comparison, although Gaussian blur is not able to remove all noisy edges, it does not introduce new ones. 

\section{Conclusion \& Future Work}\label{sec:conclusion}

For automated coastline extraction, a combination of PSNR or RSME and SSIM or UQI metrics can be used to compare edge detection algorithms. The results of these metrics produce similar conclusions when comparing the algorithms visually. Using these metrics and visual analysis, we saw that Canny produced the best results. It was the most robust to noisy non-coastal edges. Edge detection algorithms may be able to perform well for some coastlines. However, we found that even Canny was not robust to all noisy edges. Particularly, we saw it was not able to distinguish edges caused by property development on the coastline. This is an issue when developing a scalable solution that would be expected to monitor a variety of coastlines. 

Based on the findings of this paper, we suggest future research focuses on deep learning methods. Using a variety of training images, a deep learning model can learn the difference between a coastal and non-coastal edge. In other words, these models can be more robust to noisy edges. Deep learning methods may still benefit from some preprocessing done in this analysis. Particularly, histogram equalization could be used as a feature engineering method. In the future, we plan to analyze the effect of indices as an additional method. We also plan to release a similar dataset to SWED with a focus on the Irish coastline.


\balance
\bibliographystyle{IEEEtran.bst}
\bibliography{longforms,refs}

\end{document}